\documentclass[10pt,twocolumn,letterpaper]{article}

\usepackage{iccv}
\usepackage{times}
\usepackage{epsfig}
\usepackage{graphicx}
\usepackage{amsmath}
\usepackage{amssymb}
\usepackage{booktabs}
\usepackage{multirow}
\usepackage{authblk}

\usepackage[pagebackref=true,breaklinks=true,letterpaper=true,colorlinks,bookmarks=false]{hyperref}

\iccvfinalcopy 

\def\iccvPaperID{5474} 

\begin{document}

\title{Low Pass Filter for Anti-aliasing in Temporal Action Localization}

\author[$1$, $2$]{Cece Jin}
\author[$1$, $2$]{Yuanqi Chen}
\author[$1$]{Ge Li\thanks{$^{\ast}$Corresponding Author.}}
\author[$1$]{Tao Zhang}
\author[$3$]{Thomas Li}

\affil[$1$]{ School of Electronic and Computer Engineering, Peking University}
\affil[$2$]{Peng Cheng Laboratory}
\affil[$3$]{Advanced Institute of Information Technology, Peking University}
\affil[ ]{\textit {\{fordacre, cyq373, t\_zhang\}@pku.edu.cn,  geli@ece.pku.edu.cn, tli@aiit.org.cn}}

\maketitle


\begin{abstract}
In temporal action localization methods, temporal downsampling operations are widely used to extract proposal features, but they often lead to the aliasing problem, due to lacking consideration of sampling rates. This paper aims to verify the existence of aliasing in TAL methods and investigate utilizing low pass filters to solve this problem by inhibiting the high-frequency band. However, the high-frequency band usually contains large amounts of specific information, which is important for model inference. Therefore, it is necessary to make a tradeoff between anti-aliasing and reserving high-frequency information. To acquire optimal performance, this paper learns different cutoff frequencies for different instances dynamically. This design can be plugged into most existing temporal modeling programs requiring only one additional cutoff frequency parameter. Integrating low pass filters to the downsampling operations significantly improves the detection performance and achieves comparable results on THUMOS'14, ActivityNet~1.3, and Charades datasets. Experiments demonstrate that anti-aliasing with low pass filters in TAL is advantageous and efficient.
\end{abstract}

\section{Introduction}
\label{introduction}
\begin{figure}[t]
\vskip 0.2in
\begin{center}
\centerline{\includegraphics[width=\columnwidth]{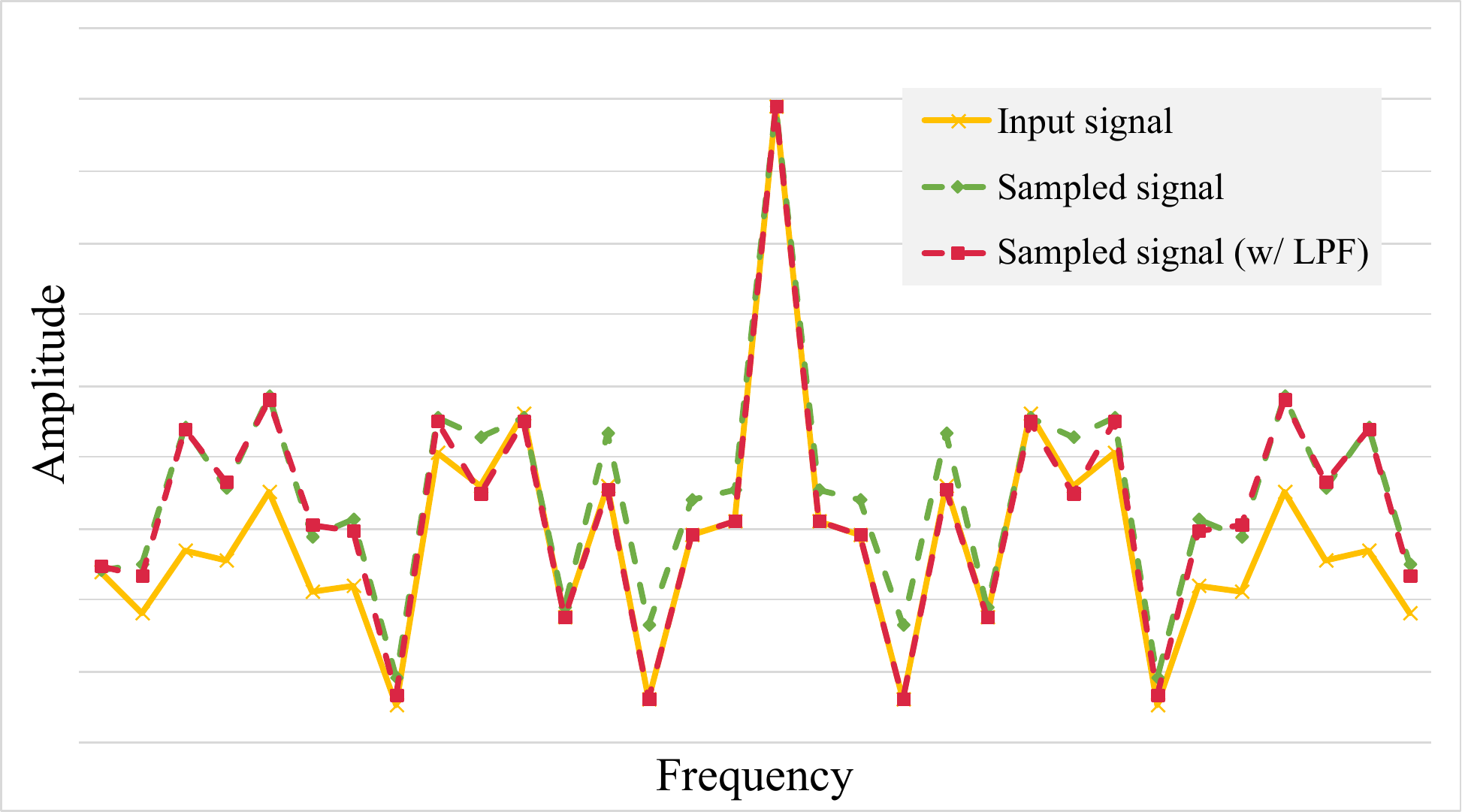}}
\vskip 0.1in
\caption{A manifestation of the aliasing problem in the frequency domain. Due to the aliasing, the sampled feature has a very different spectrum compared to the original feature. Inserting an additional low pass filter significantly reduces the aliasing error at the low-frequency band.}
\label{fig:intro}
\end{center}
\vskip -0.4in
\end{figure}

In temporal action localization (TAL), temporal downsampling operations, such as pooling and frame sampling, are widely used to reshape proposals with variable temporal lengths into a fixed size. Nevertheless, these simple downsampling methods neglect considering the relationship between the sampling rate and the highest frequency of the original input. They often employ a low sampling rate which would lead to the well-known aliasing problem. As shown in Figure~\ref{fig:intro}, the most common manifestation of the aliasing problem is that the frequency spectrum of the sampled feature differs from the original feature. This practice will cause features sampled from different sources indistinguishable and the sampled feature can't restore to the original one (for example, the wagon-wheel effect) \cite{Mitchell1988}. That means the sampled feature might lose some important information of the original feature and the features of two different actions might be indistinguishable after a temporal downsampling operation. This paper investigates eliminating the aliasing error by integrating low pass filters to temporal downsampling operations.

The Low Pass Filter (LPF) can efficiently attenuate the high-frequency information and retain the low-frequency information. Therefore, it is widely used for anti-aliasing in digital signal processing and image processing. According to the Fourier Transform, setting the cutoff frequency of the LPF to $\pi$ divides the downsampling factor can eliminate the aliasing error in maximum. Under this circumstance, the LPF completely removes the exceeding high-frequency part, which is the chief culprit of the aliasing. On the other hand, Wang \etal \cite{Wang2020} proposes that the high-frequency band contains abundant personalized information that is important for model inference. Therefore, cutting the high-frequency part might actually counterproductively hurt the detection performance. It is necessary to find an equilibrium point between removing aliasing and reserving high-frequency information by adjusting the cutoff frequency. To solve this problem, several convolution layers were used to learn the cutoff frequencies of different proposals dynamically in this paper. This method allows the network to determine the optimal points for each proposal to balance the influence of anti-aliasing and retaining high-frequency information. Additionally, this method can be easily applied to substantially all downsampling methods at barely any additional costs. To validate the effectiveness of anti-aliasing with low pass filters, extensive experiments were conducted on the THUMOS'14 \cite{Jiang2014} dataset, the ActivityNet 1.3 \cite{caba2015activitynet} dataset, and the Charades \cite{Sigurdsson2016} dataset by integrating low pass filters to pooling, frame sampling, and convolution kernels respectively.

There are several contributions in this paper:
(1) Proposing to alleviate the aliasing problem with low pass filters in TAL.
(2) Proposing to adjust cut-off frequencies dynamically to balance the tradeoff between anti-aliasing and reserving high-frequency information.
(3) Achieving comparable performances on THUMOS'14, ActivityNet 1.3, and Charades datasets.

\section{Related Works}
Video analysis is an important aspect of computer vision and machine learning. In the past decades, plenty of works have been published in this field and greatly promoted its development. In the beginning, \cite{Laptev2003, Dalal2005, Wang2013} carried out hand-crafted features (such as STIPs, HOG, iDT) to represent videos. Influenced by the success of the convolution network in image classification, \cite{Simonyan2014} put forward Two Stream Convolution Networks utilizing 2D convolution to capture spatial and temporal information independently. In the next year, C3D \cite{Tran2015} conducted a series of experiments that use 3D convolution to learn features of spatiotemporal inputs directly. Most of the subsequent works adopt both two-stream and 3-dimensional convolution at the same time to recognize actions. I3D inflated off-the-shelf ImageNet pretrained 2D convolution network into 3D and improved the recognition performance significantly. Moreover, Non-local \cite{Wang2018} achieved a remarkable result by utilizing the popular attention mechanism to capture long-range dependencies.

However, most classification approaches suffer from the fatal drawback of being incapable of managing untrimmed long videos. In temporal action detection, works stacked different neural networks on top of the aforementioned recognition methods to make up for this deficiency. DAPs \cite{Escorcia2016} and SST \cite{Buch2017} use RNN to yield proposal information, such as activity category and location, every timestep. Most other works followed a local-to-global fashion that they used 1D temporal convolution to learn local relationships and frame sampling \cite{Dai2017, Lin2018, Gao*2018, Lin2019, Lin2020, Xu2020}, average pooling \cite{Gao2017}, max pooling \cite{Zeng2019}, or pyramid pooling variants \cite{Xu2017, Zhao2017, Chao2018, Li2020} to extract global proposal features. Different from previously mentioned methods, SSAD \cite{Lin2017}, Decouple-SSAD \cite{Huang2019}, and PBRnet\cite{Liu2020} made predictions on the anchor layers using 1D temporal convolution layers. GTAN \cite{Long2019} further proposed Gaussian Pooling to adjust the anchors and their receptive fields. There are also methods maintained employing convolution-deconvolution manner for the video segmentation problem. In \cite{Shou2017}, convolution and deconvolution are incorporated into one CDC filter which upsamples in the time dimension while downsampling in the spatial dimension. To manage varied action temporal lengths, TCNs \cite{Lea2017} and TDRN \cite{Lei2018} utilize dilated convolution and deformable convolution \cite{Dai2017a} to adapt the temporal receptive field to a suitable size. Temporal Gaussian Mixture \cite{Piergiovanni2019a} modified the traditional convolution operation by learning several Gaussian distributed kernel weights to capture long-term dependencies in a full temporal resolution. However, the downsampling mechanism, no matter pooling or uniform sampling, used in these methods often leads to the aliasing problem. The convolution kernel is also a kind of downsampling and suffers from aliasing. To alleviate the negative effect of downsampling, low pass filters are utilized to inhibit the aliasing error in this paper.

Recently, Wang \etal \cite{Wang2020} found that the CNN is encouraged to capture high-frequency information that is not perceivable to humans to improve classification performance. Zhang \cite{Zhang2019} utilized low pass filters to make popular convolution neural networks shift-invariant in the 2D spatial dimension. Different from \cite{Zhang2019}, this work focuses on using low pass filters to alleviate the aliasing problem caused by downsampling in the 1D temporal dimension. Additionally, the balancing the tradeoff between anti-aliasing and reserving high-frequency information to acquire optimal detection performance is investigated.

\section{Approach}
This section will introduce the Low Pass Filter and discuss why conventional temporal downsampling methods, such as uniform sampling, would lead to aliasing. The tradeoff between using low pass filters for anti-aliasing and increasing the cutoff frequencies to reserve more high-frequency information will also be discussed.
\subsection{Low Pass Filter}
\label{sec:lpa}

The Low Pass Filter (LPF) is a simple and effective filter that is widely used in communication systems to remove redundant information in the frequency domain. The ideal LPF is a rectangular window in the frequency domain that blocks signals with frequencies higher than the cutoff frequency while passing the signals with frequencies lower than the cutoff frequency. The LPF can be represented in the frequency domain as:
\begin{equation}
H(\omega) = \left\{
\begin{aligned}
1 & , & \vert \omega \vert \leq f_C, \\
0 & , & \vert \omega \vert > f_C,
\end{aligned}
\right.
\end{equation}
where $\omega$ is the frequency variable and $f_C$ represents the cutoff frequency. The LPF is usually used to attenuate the high-frequency band by multiplying the low pass function with the input signal $X(\omega)$ in the frequency domain:
\begin{equation}
Y(\omega)= H(\omega) \cdot X(\omega) = \left\{
\begin{aligned}
X(\omega) & , & \vert \omega \vert \leq f_C, \\
0,\ \ \quad & & \vert \omega \vert > f_C.
\end{aligned}
\right.
\end{equation}
However, signals are represented as $x(t)$ using time base in most cases and it is costly to convert them into the frequency domain. Therefore, converting the LPF into the time domain using the Fourier Transformation would work better (refer to supplementary materials for more details). Thus, the filtered signal can be represented in the time domain as:
\begin{equation}
y(t) = h(t) \otimes x(t) = Sinc(f_C \cdot t) \otimes x(t).
\label{eqa:lpf_t}
\end{equation}
Moreover, assume we have discrete input $x(n)$ in practice, the output can be represented as:
\begin{equation}
y(n) = Sinc(f_C \cdot n) \otimes x(n).
\label{eqa:lpf_n}
\end{equation}
Constant terms have been omitted for convenience. In this paper, signals refer to temporal features.

\subsection{Downsampling and Anti-aliasing}
\label{sec:aliasing}
\begin{figure}[t]
\vskip 0.1in
\begin{center}
\centerline{\includegraphics[width=\columnwidth]{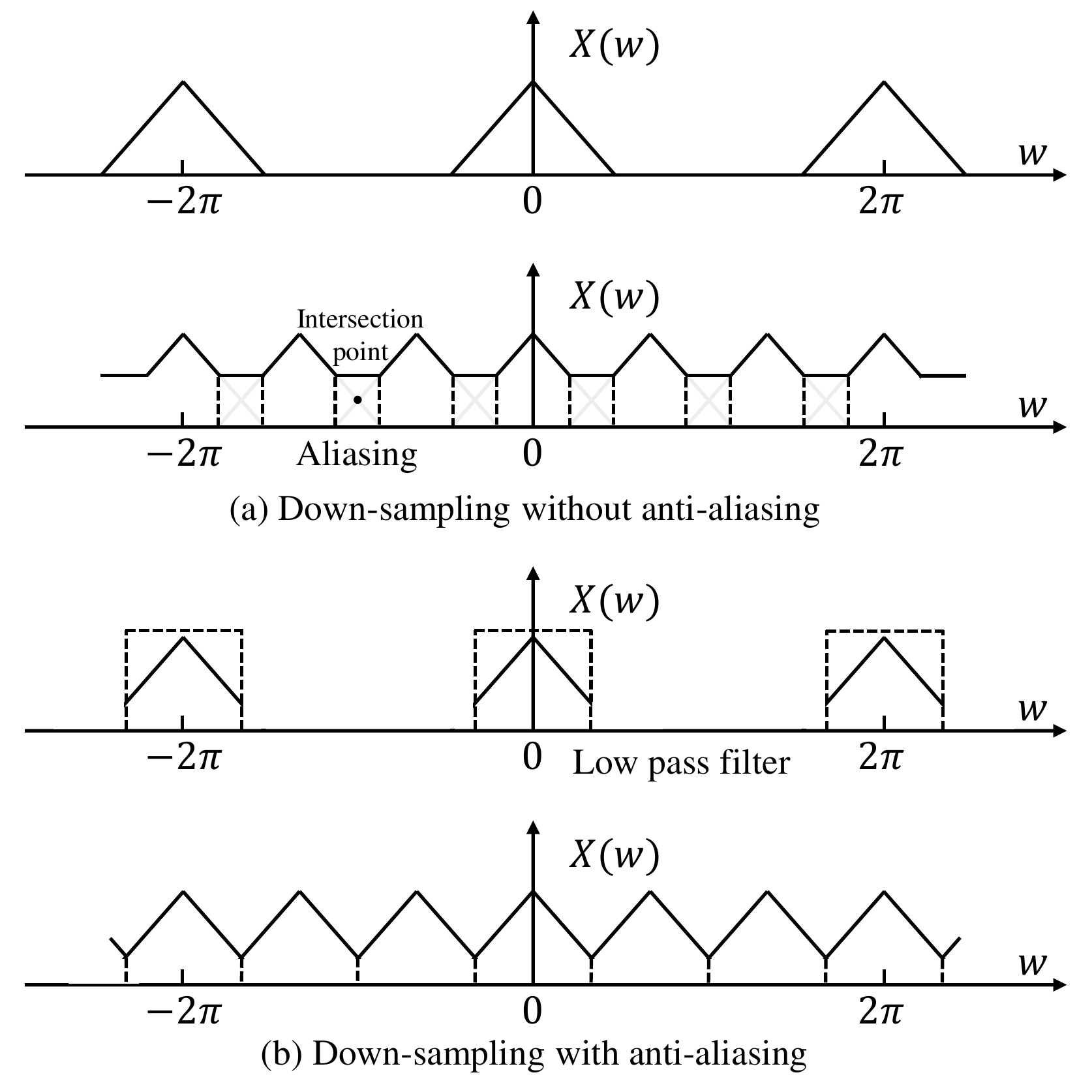}}
\caption{Downsampling and Anti-aliasing. (a) When downsampling a signal with an inadequate sampling rate, the aliasing problem distorts the frequency spectrum of the original signal. (b) Aliasing can be mitigated by integrating a low pass filter before the downsampling. The low-pass filter attenuate high-frequency band and thus inhibits high-frequency band aliasing.}
\label{fig:aliasing}
\end{center}
\vskip -0.4in
\end{figure}

For a signal $x(n)$, we downsampling it by an integer factor $N$:
\begin{equation}
x_p(n) = x(n) \cdot p(n) = \sum_{k=-\infty}^{\infty} x(kN) \delta(n-kN),
\end{equation}
where the sampling function
\begin{equation}
p(n) = \sum_{k=-\infty}^{\infty} \delta(n-kN),
\end{equation}
and $\delta$ represent the impulse function.
Transforming $x_p(n)$ into frequency domain results in:
\begin{equation}
X_p(\omega) = X(\omega) \otimes P(\omega) = \sum_{k=0}^{N-1} X(\omega-\frac{2\pi}{N}k),
\label{eqa:sampling}
\end{equation}
where
\begin{equation}
P(\omega) = \sum_{k=-\infty}^{\infty} \delta(\omega-\frac{2\pi}{N}k).
\end{equation}
Due to convolution operation in the frequency domain (see Equation~\ref{eqa:sampling}), the exceedingly high-frequency band could be mixed up and form a new shape which causes aliasing (see Figure~\ref{fig:aliasing} (a)). As thus the sampled signal may lose some important information and cannot restore to the original reference. When equipped with an ideal LPF (see Figure~\ref{fig:aliasing} (b)), the high-frequency band of the original signal is attenuated. This inhibits the aliasing of high-frequencies from spreading into the lower frequency band and destroying the original modality. With the LPF, the frequency spectrum can preserve the original waveform more accurately. Although the LPF effectively removes the aliasing error, it pays a price in losing the high-frequency band higher than the intersection point. On the other hand, as the cutoff frequency increases, more high-frequency information can be reserved while the anti-aliasing effect decreases. Therefore, adjusting the cutoff frequency dynamically to balance the effects of anti-aliasing and reserving high-frequency information may behave better. During the experiment, low pass filters are inserted before the temporal downsampling operations, including pooling, frame sampling, and convolution kernel, to improve the detection performance. Models are set to learn different cutoff frequencies for different instances to achieve better performance. Relevant details can be found in the following experiments section.

\section{Experiments}
We conduct extensive experiments on THUMOS'14, ActivityNet 1.3, and Charades datasets by inserting the LPF into pooling, frame sampling, and convolution kernels respectively. In practice, most temporal action localization models (especially the two-stage methods) take advantage of frame sampling or pooling to align and extract proposal features. By conducting experiments on frame sampling and pooling, one can validate the broad applicability of anti-aliasing with LPF in temporal action localization. Some traditional one-stage methods might not have such downsampling methods as they make predictions on anchor layers directly. 
However, GTAN \cite{Long2019} proved the superiority of utilizing Gaussian pooling to adjust anchors' sizes and receptive fields for one-stage temporal action localization methods. Therefore, this paper follows the manner of GTAN and inserts pooling operations into one-stage methods. 
This process allows the validation of the effectiveness of anti-aliasing with LPF in one-stage methods.

\subsection{Low Pass Filter + Pooling}
\label{section:pooling}
To validate the effectiveness of anti-aliasing for pooling operations, LPF is integrated into the classic average pooling in this section. Assume an input sequence of $x(n)$ with the length of $N$. According to Equation~\ref{eqa:lpf_n}, one can define the LPF + average pooling as:
\begin{equation}
P^{lp} = \frac{1}{N} \sum_{n=1}^N [Sinc(f_C \cdot n) \otimes x(n)].
\label{eqa:lpp}
\end{equation}
Additionally, simplify the Equation~\ref{eqa:lpp} to save the computation overhead as:
\begin{equation}
P^{lp} = \frac{1}{Z} \sum_{n=1}^N \sum_{m=1}^{N}Sinc(f_C(n - m - \mu)) \cdot x(n),
\end{equation}
where $Z$ is the normalization factor, and $\mu$ is the shift constant. The detailed deduction can be seen in the supplementary materials. For convenience, the equation
\begin{equation}
w^{lp}_n = \frac{1}{Z} \sum_{n=1}^N \sum_{m=1}^{N}Sinc(f_C(n - m -\mu))
\label{eqa:lpw}
\end{equation}
is defined as the low pass weight.

\textbf{Implementation details.}
For LPF + pooling, this paper takes the SSAD \cite{Lin2017} with the Gaussian Kernel proposed by the GTAN \cite{Long2019} as a baseline. Frozen I3D is used as a feature extractor. Extracted rgb and flow features are early fused using a convolution layer with kernel size 3 and stride 1. To better capture long-term temporal information, the number of anchor layers is expanded from 3 to 6 and the anchor aspect ratio is squeezed from \{0.5, 0.75, 1.0, 1.5, 2.0\} to \{1.0\}. The Adam optimizer is used to train the model for 30 epochs with a batch size of 16 and a learning rate of 0.0001. The LPF + pooling layer is inserted between the anchor layer and the prediction layer to extract proposals' features. Cut-off frequencies are learnt using a conv1d(channel\_in, anchor\_num, 3) (conv1d(in\_channel, out\_channel, kernel\_size)) layer. It takes anchor layer feature $x(n)$ with shape (B, T, C) as inputs and outputs results with shape (B, T, anchor\_num), where B, C, T represent batch\_size, channel\_num, and time\_len respectively. That means one proposal corresponds to one LPF.

\textbf{Dataset.}
THUMOS'14 is a widely used dataset in video analysis. For the temporal action detection task, there are 200 untrimmed videos of the validation set and 213 videos of the testing set annotated with locations, and 20 action classes. In this paper, models are trained on the validation set and evaluated on the testing set. All experiment results are evaluated using the official EvalKit of THUMOS'14.

\subsubsection{Quantitative analysis}
\begin{table}[t]
\caption{Comparison between different pooling methods on temporal action localization task on THUMOS'14 (mAP@0.5).}
\label{tab:pooling}
\vskip 0.15in
\begin{center}
\begin{small}
\begin{tabular}{lcccr}
\toprule
Method & Detection Result (\%) \\
\midrule
Average Pooling & 41.87\\
Max Pooling & 44.69\\
Softmax Attention Pooling & 45.34\\
Sigmoid Attention Pooling & 45.90\\
Gaussian Pooling \cite{Long2019} & 46.29\\
\midrule
Gaussian LPF + Average Pooling & 49.50\\
LPF + Average Pooling & \textbf{49.89}\\
\bottomrule
\end{tabular}
\end{small}
\end{center}
\vskip -0.2in
\end{table}
This paper compares the different pooling methods based on the baseline model, especially for pooling methods with and without LPF. As shown in Table~\ref{tab:pooling}, the LPF + average pooling outperforms the naive average pooling by 8.02\%. This demonstrates that using the LPF to reduce the aliasing error can improve the performance greatly. The Sinc function of the LPF is replaced with Gaussian function as Gaussian LPF  for ablation study. The Gaussian LPF + average pooling achieves 49.50\% mAP@0.5 which outperforms the average pooling by 7.65\% and shows a minor 0.39\% decrease compared to the Sinc LPF. This indicates that the form of the low pass filter is not the key factor of anti-aliasing as long as it can inhibit the high-frequency band efficiently. Table~\ref{tab:pooling} compares the LPF + average pooling to other pooling methods. Thereinto, the sigmoid attention pooling, softmax attention pooling, and Gaussian pooling represent pooling operations with sigmoid, softmax, and Gaussian attention assigned to the pooling kernels respectively. LPF + average pooling achieves a remarkable 49.89\% of mAP@0.5 and outperforms all other pooling methods by a large margin. These results suggest using LPF for anti-aliasing in TAL.

\begin{table*}[t]
\caption{Temporal action localization mAP (\%) on THUMOS'14.}
\label{tab:thumos}
\vskip 0.1in
\begin{center}
\begin{small}
\begin{tabular}{lcccccccc}
\toprule
tIoU& 0.1& 0.2& 0.3& 0.4& 0.5& 0.6 & 0.7\\
\midrule
BSN \cite{Lin2018}& -& -& 53.5& 45.0& 36.9& 28.4& 20.0\\
TAL-Net \cite{Chao2018}& 59.8& 57.1& 53.2& 48.5& 42.8& 33.8& 20.8\\
GTAN \cite{Long2019}& 69.1& 63.7& 57.8& 47.2& 38.8& -& -\\
PGCN \cite{Zeng2019}& 69.5 & 67.8 & 63.6 & 57.8 & 49.1 &- &-\\
TGM \cite{Piergiovanni2019a} & - & -& -& -& \textbf{57.1}& - & -\\
G-TAD \cite{Xu2020} + PGCN & - & -& 66.4& 60.4& 51.6& 37.6 & 22.9\\
PBRNet \cite{Liu2020} & - & -& 58.5& 54.6& 51.3& \textbf{41.8} & \textbf{29.5}\\
AGCN \cite{Li2020} & 59.3 & 59.6& 57.1& 51.6& 38.6& 28.9 & 17.0\\
\midrule
SSAD \cite{Lin2017} (I3D backbone) & 63.3& 62.0& 59.4& 54.2& 44.4& 30.2& 15.7\\
SSAD + LPF (I3D backbone) & 64.5& 63.1& 60.3& 55.2& 46.0& 32.1& 15.7\\
\midrule
G-TAD \cite{Xu2020} (TSN \cite{Wang2016} backbone) & - & -& 54.5 & 47.6 & 40.2 & 30.8 & 23.4\\
G-TAD + LPF (TSN backbone) & - & - & 56.9 & 51.2 & 42.7 & 32.9 & 22.8\\
\midrule
Ours & \textbf{71.8} & \textbf{69.3}& \textbf{67.0}& \textbf{61.7}& 52.6& 38.9 & 23.0\\
\bottomrule
\end{tabular}
\end{small}
\end{center}
\vskip -0.35in
\end{table*}

\subsubsection{Qualitative analysis}
\begin{figure}[t]
\vskip 0.1in
\begin{center}
\begin{minipage}[t]{0.49\linewidth}
\centering
\centerline{\includegraphics[width=1\linewidth]{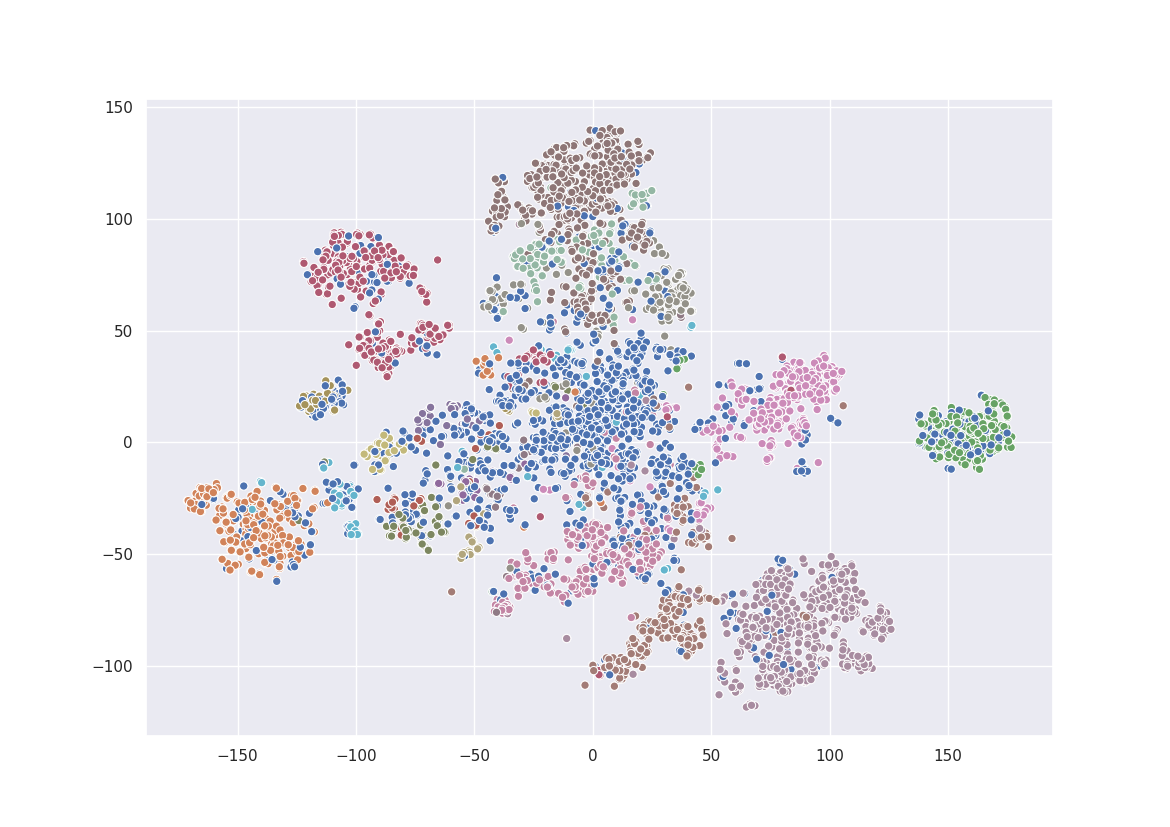}}
\small
\centerline{(a) Layer 3: w/o LPF}\medskip
\end{minipage}
\hfill
\begin{minipage}[t]{0.49\linewidth}
\centering
\centerline{\includegraphics[width=1\linewidth]{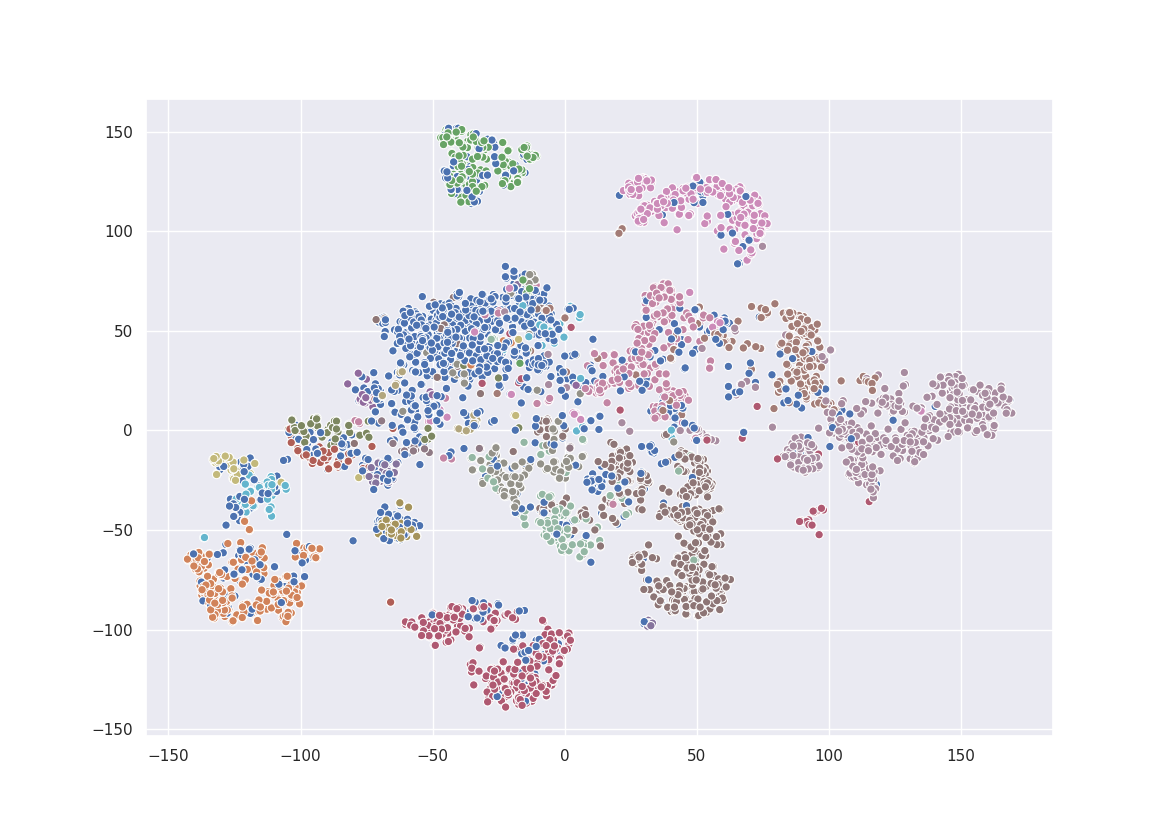}}
\small
\centerline{(b) Layer 3: w/ LPF}\medskip
\end{minipage}
\hfill
\vskip -0.1in
\begin{minipage}[t]{0.49\linewidth}
\centering
\centerline{\includegraphics[width= 1\linewidth]{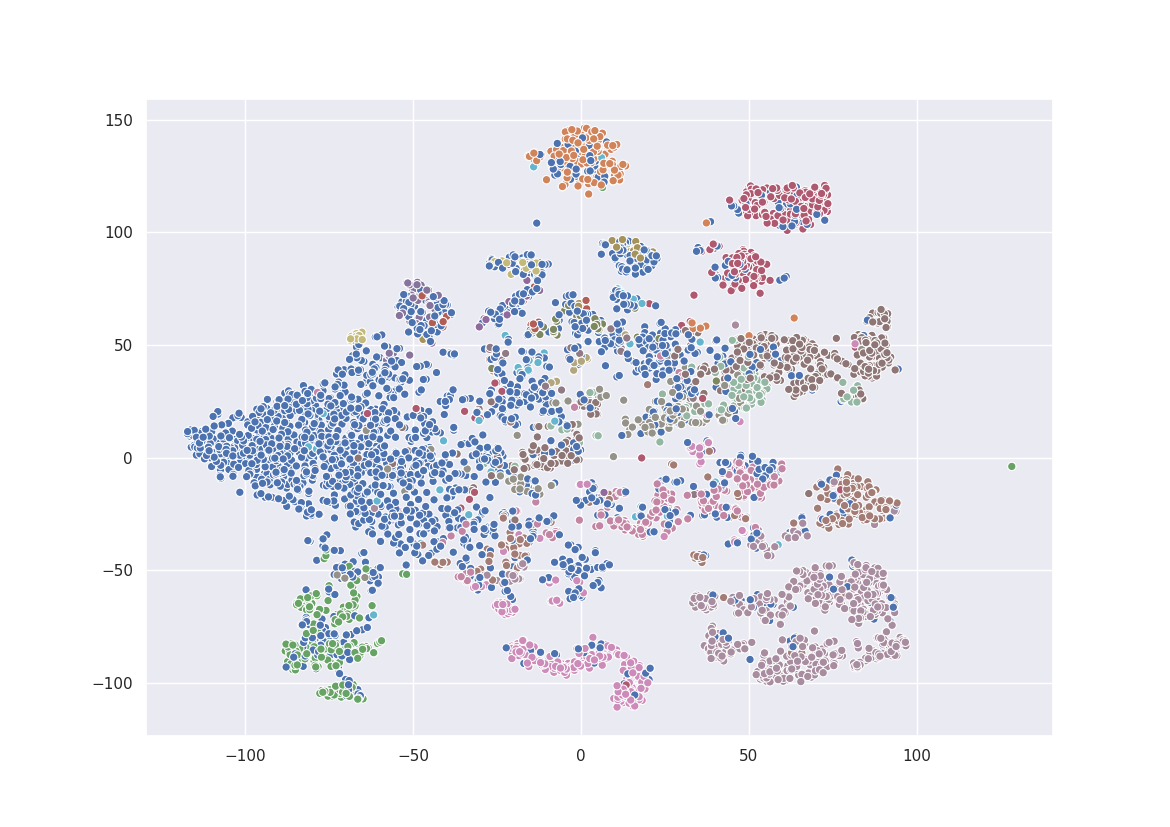}}
\small
\centerline{(c) Layer 4: w/o LPF}\medskip
\end{minipage}
\hfill
\begin{minipage}[t]{0.49\linewidth}
\centering
\centerline{\includegraphics[width=1\linewidth]{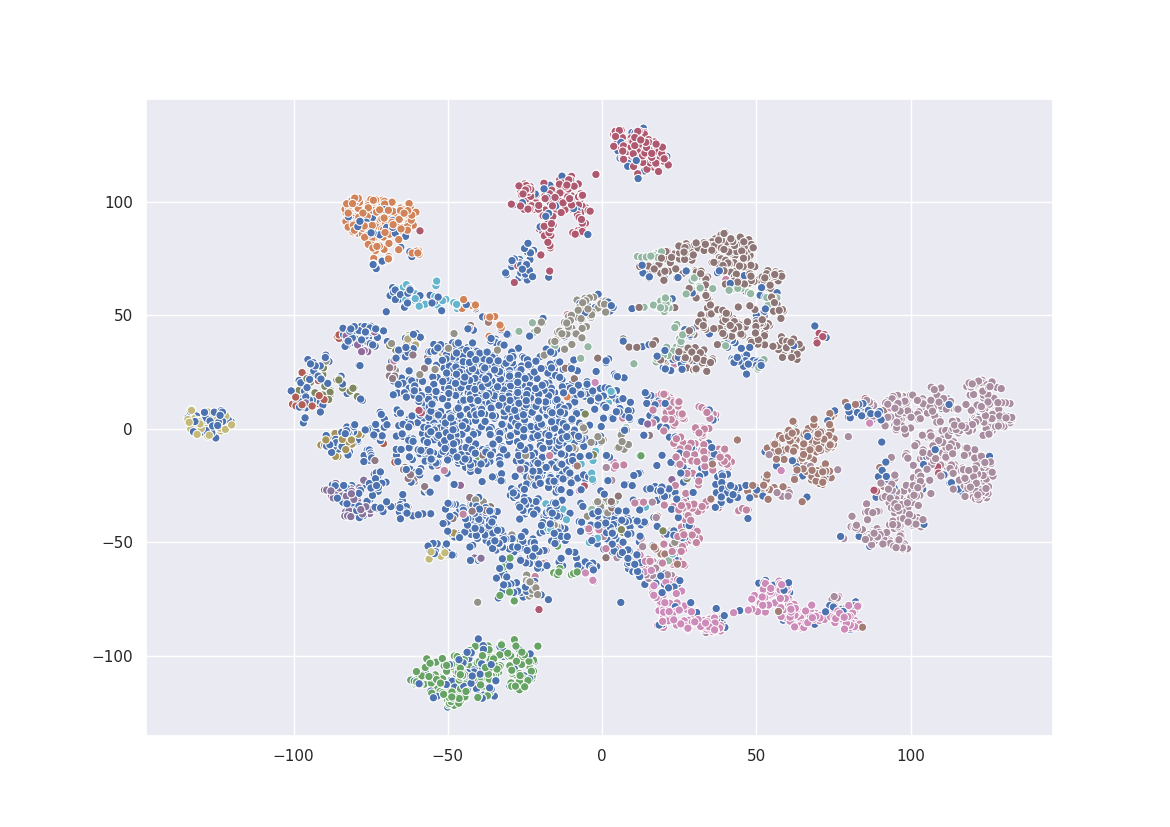}}
\small
\centerline{(d) Layer 4: w/ LPF}\medskip
\end{minipage}
\vskip 0.1in
\caption{Visualization of proposals' features at different anchor layers using the t-SNE. The major blue points correspond to the background. Part of the background proposals has been filtered out randomly for clarity.}
\label{fig:tsne}
\end{center}
\vskip -0.25in
\end{figure}

\begin{figure}[t]
\vskip 0.2in
\begin{center}
\centerline{\includegraphics[width=\columnwidth]{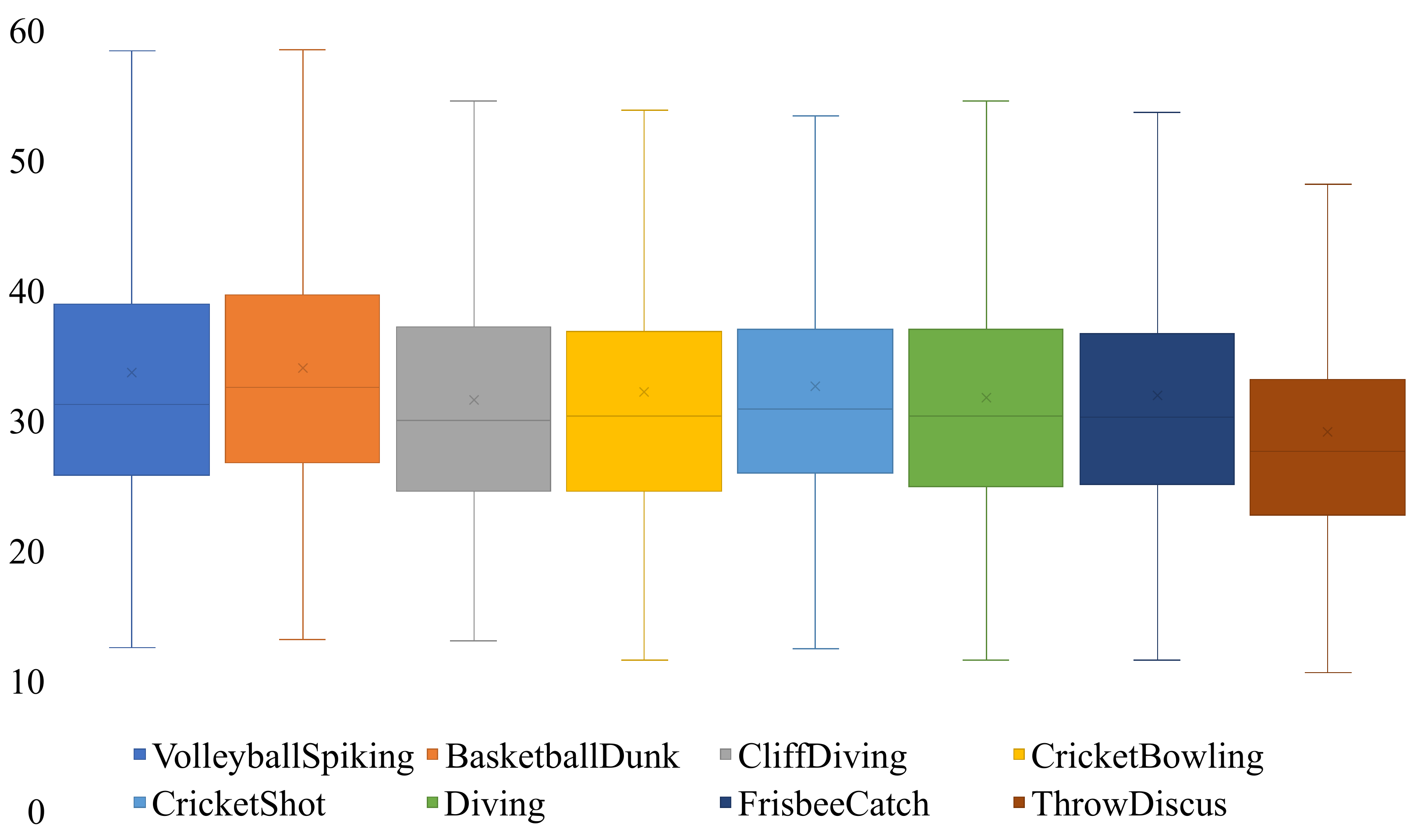}}
\vskip 0.1in
\caption{The cutoff frequencies of low pass filters for different action instances. Eight action classes are randomly selected from all twenty classes.}
\label{fig:freqs}
\end{center}
\vskip -0.4in
\end{figure}

As can be seen from Figure~\ref{fig:tsne}, the proposals' features extracted with LPF show better separability than without LPF. This demonstrates that removing the aliasing error can provide more differentiated features and promote classification performance. A collection of the cutoff frequencies of different action instances are shown in Figure~\ref{fig:freqs}. The box figure shows that different action instances have different cutoff frequencies $f_C$ and different action classes have different average values of $f_C$. This verifies that there exists a dynamic game between anti-aliasing and reserving high-frequency information. Average cutoff frequencies of vigorous actions (such as BasketballDunk) are noticeable higher than moderate actions (such as ThrowDicus). For dunking a basketball, the high-frequency band contains more valuable information compared to throwing a discus. Consequently, the neural network chooses a higher cutoff frequency as reserving high-frequency information is more rewarding than anti-aliasing. This indicates the model tries to optimize the detection performance by adjusting the cutoff frequency.

\subsubsection{Comparison with state-of-the-art methods}
This subsection compares our method to several state-of-the-art methods. Table~\ref{tab:thumos} presents the result evaluated using mAP values with different tIoU thresholds. Our method, with the I3D backbone, achieves 52.6\% mAP@0.5, outperforming one-stage model PBRNet by 1.3\% and two-stage model G-TAD by 1.0\%. Our method also surpasses other methods at tIoU thresholds [0.1, 0.2, 0.3, 0.4], proving that anti-aliasing by inserting low pass filters before downsampling operations is advantageous. Additionally, LPF is applied to the SSAD and G-TAD to prove its broad adaptability. Unsurprisingly, using LPF to alleviate the aliasing problem in SSAD and G-TAD improves the performance by 1.6\% and 2.5\% of mAP@0.5 respectively.

\subsection{Low Pass Filter + Frame Sampling}
\label{section:sampling}
\begin{figure}[t]
\vskip 0.1in
\begin{center}
\begin{minipage}[b]{0.49\linewidth}
\centering
\centerline{\includegraphics[width=\linewidth]{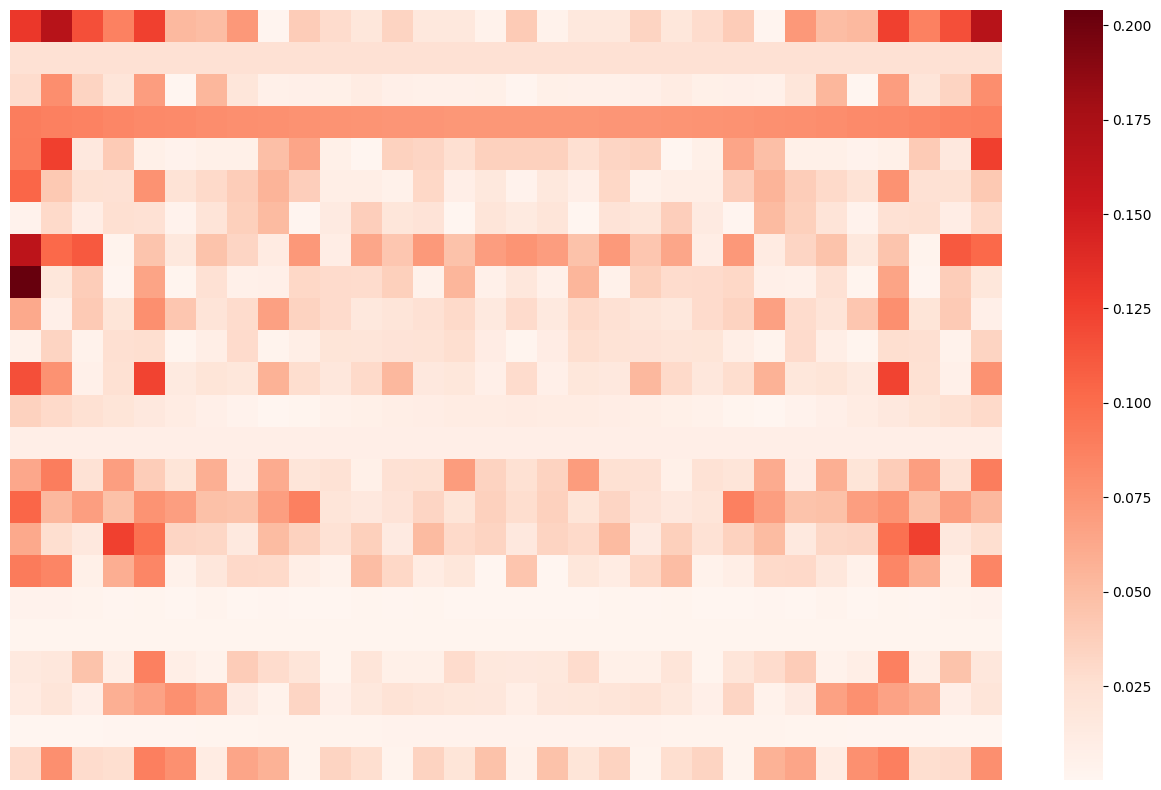}}
\small
\centerline{(a) w/o LPF}\medskip
\end{minipage}
\hfill
\begin{minipage}[b]{0.49\linewidth}
\centering
\centerline{\includegraphics[width= \linewidth]{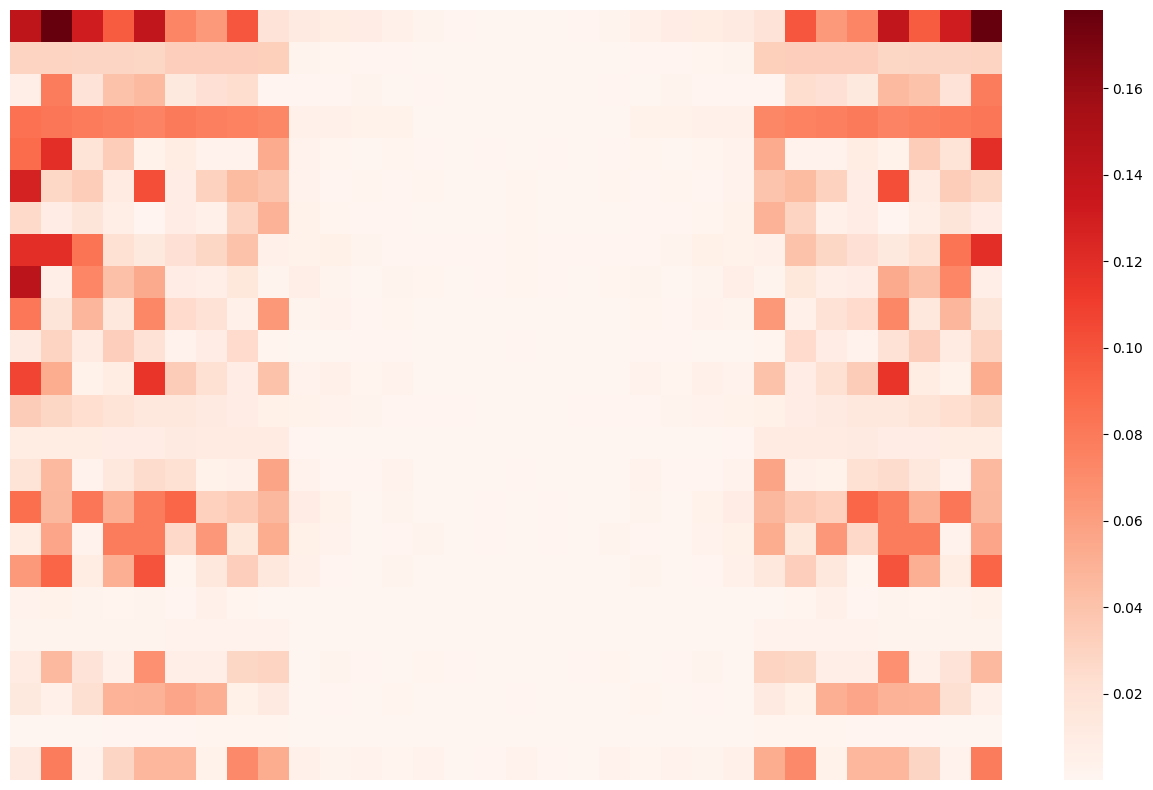}}
\small
\centerline{(b) w/ LPF}\medskip
\end{minipage}
\caption{Difference between the uniformly sampled feature and the original feature in the frequency domain. The vertical axis represents channel. The horizontal axis represents frequency.}
\label{fig:heat}
\end{center}
\vskip -0.2in
\end{figure}

Frame sampling is widely used for temporal downsampling in TAL. This section validates the effectiveness of combining LPF with uniform sampling for anti-aliasing. It can be defined as:
\begin{equation}
S^{lp}= \{Sinc(f_C \cdot n) \otimes x(n)\}_{\ n\ mod\ k\ =\ 0},
\label{eqa:slp}
\end{equation}
where $k$ represents the downsampling factor.

\textbf{Implementation details.} This section takes the well-performed temporal action proposal model BMN \cite{Lin2019} as the baseline. Because frame sampling is used in the Boundary-Matching module to construct proposal features, low pass filters are placed before the Boundary-Matching module for anti-aliasing. The sampling method in BMN is changed into uniform sampling as shown in Equation~\ref{eqa:slp} while other settings are kept the same. Cut-off frequencies are learnt using cnn block (conv1d(256, 256, 3), relu, avgpool1d(100), conv1d(256, 1, 1), sigmoid) (avgpool1d(kernel\_size)). It takes the previous temporal feature sequence $x(n)$ with shape (B, C, T) as inputs and outputs results with shape (B, ). That means one video clip corresponds to one LPF.

\textbf{Dataset.} The ActivityNet 1.3 dataset consists of over 20000 untrimmed videos covering 200 daily living activity classes. Specifically, 10024 videos are of the training subset, 4926 videos are of the validation subset, and 5044 videos are of the testing subset. For the ActivityNet dataset, most of the experiments are conducted on the temporal action proposal task and take the AR@AN value as the evaluation metric. Models are trained on the training set and evaluated on the validation set. All experiment results are evaluated using the official EvalKit of ActivityNet 1.3.

\begin{table}[t]
\caption{Comparison between uniform sampling w/ and w/o the LPF with different sampling rates (AUC(\%)).}
\label{tab:sample}
\vskip 0.15in
\begin{center}
\begin{small}
\begin{tabular}{ccccr}
\toprule
Samples number & w/o & w/ & \\
\midrule
2 & 65.92 & \textbf{66.27} & +0.35\\
3 & 66.30 & \textbf{66.56} & +0.26\\
4 & 66.65 & \textbf{66.89} & +0.24\\
8 & 67.07 & \textbf{67.31} & +0.24\\
16 & 67.34 & \textbf{67.41} & +0.07\\
32 & 67.43 & \textbf{67.53} & +0.10\\
\bottomrule
\end{tabular}
\end{small}
\end{center}
\vskip -0.15in
\end{table}

\begin{table}[t]
\caption{Comparison between different cutoff settings for LPF + uniform sampling with 2 samples (AUC(\%)). In this example, cutoff equals 50 means without LPF.}
\label{tab:cutoff}
\vskip 0.15in
\begin{center}
\begin{small}
\begin{tabular}{lcccccc}
\toprule
Cutoff & 10 & 20 & 30 & 40 & 50 & Ours \\
\midrule
AUC & 65.68 & 65.91 & 66.07 & 66.15 & 65.92 & \textbf{66.27} \\
\bottomrule
\end{tabular}
\end{small}
\end{center}
\vskip -0.2in
\end{table}

\textbf{The existence of aliasing and the effectiveness of LPF for anti-aliasing.} As shown in Figure~\ref{fig:heat} (a), frequency differences are plotted between the sampled feature and the original feature in order to observe the aliasing problem clearly. It is evident that the sampled feature different from the original feature across the entire frequency domain. This validates the existence of the aliasing and means the messages contained in the original feature will be very different after the downsampling operation. Integrating LPF to uniform sampling efficiently suppresses the aliasing error in the low-frequency band (see Figure~\ref{fig:heat} (b), the empty space in the middle of the frequency spectrum), which validates that the LPF is effective for anti-aliasing. It is worth mentioning that the model learns an eclectic cutoff frequency that doesn't wipe out the aliasing completely (see Figure~\ref{fig:heat} (b), the red residual in both sides of the frequency spectrum) but chooses to reserve some high-frequency information. This result supports the proposition of balancing the influence between anti-aliasing and retaining high-frequency information.

\textbf{The utility of anti-aliasing with LPF in TAL.} As can be seen in Table~\ref{tab:sample}, adding low pass filters to frame sampling operations can consistently improve the performance with sample numbers from 2 to 32. This shows that suppressing the aliasing problem in the low-frequency band is beneficial. However, when the sample number increases, the improvements decrease and provide only a 0.1\% improvement with 32 samples. This is because as the sampling number increases, the sampling rate is also increasing. Considering the aliasing degree is inversely proportional to the sampling rate, the low pass filter naturally shows minor improvement by fixing the aliasing problem. In extreme cases, when the sample number is greater than the proposal's length, the downsampling will turn into up-sampling, and aliasing does not happen anymore. On the other hand, inserting the LPF to uniform sampling with 2 samples make a 0.35\% improvement. That means in some cases when one can barely afford a low sampling rate, LPF is an appropriate choice to enhance the performance.

\textbf{The necessity of adaptive cutoff frequency.} As shown in Table~\ref{tab:cutoff}, different cutoff frequencies achieve different proposal performances. This demonstrates that the cutoff frequency has important influences on the effectiveness of anti-aliasing and the effectiveness of reserving useful high-frequency information, and it is a key factor of the model performance. Adaptive cutoff frequency achieves the best performance 66.27\% AUC compared to other manual setting cutoff frequencies. This validates that adjusting the cutoff frequencies depending on different videos to balance the relationship between anti-aliasing and reserving high-frequency information is beneficial to proposal performance.

\textbf{Comparison with state-of-the-art methods and detection results.} As seen in Table~\ref{tab:anet}, BMN + LPF achieves a comparable result, with AR@100 75.49\% and AUC 67.53\%, compared to state-of-the-art works. LPF is also applied to G-TAD for temporal action localization on the ActivityNet dataset. LPF is inserted before the SGAlign module as it utilizes sampling to align the feature. Other settings remain the same as the G-TAD paper. Cut-off frequencies are learnt using a cnn block (conv1d(256, 256, 3), relu, avgpool1d(100), conv1d(256, 256, 1), sigmoid). The block uses the GCNeXt feature with shape (B, C, T) as inputs and produces results with shape (B, C, 1). As can be seen from Table~\ref{tab:anet-det}, G-TAD + LPF outperforms the naive G-TAD 0.37\% average mAP and achieves comparable results.

\begin{table}[t]
\caption{Comparison with other works on ActivityNet 1.3(\%).}
\label{tab:anet}
\vskip 0.15in
\begin{center}
\begin{small}
\begin{tabular}{lcccccccc}
\toprule
Method & CTAP & BSN & DBG & BMN & BMN + LPF \\
\midrule
AR@100 & 73.17 & 74.16 & 76.65 & 75.01 & 75.49\\
AUC & 65.72 & 66.17 & 68.23 & 67.10 & 67.53\\
\bottomrule
\end{tabular}
\end{small}
\end{center}
\vskip -0.2in
\end{table}

\begin{table}[t]
\caption{Temporal action localization results on ActivityNet 1.3 (mAP (\%)).}
\label{tab:anet-det}
\vskip 0.15in
\begin{center}
\begin{small}
\begin{tabular}{lcccc}
\toprule
tIoU & 0.5& 0.75& 0.95& Average\\
\midrule
BMN \cite{Lin2019} & 50.07 & 34.78 & 8.29 & 33.85\\
GTAN \cite{Long2019} & 52.61 & 34.14 & 8.91 & 34.31\\
G-TAD \cite{Xu2020} & 50.36 & 34.60 & \textbf{9.02} & 34.09\\
PBRNet \cite{Liu2020} & \textbf{53.96} & 34.97 & 8.98 & \textbf{35.01}\\
\midrule
G-TAD + LPF & 50.66 & \textbf{35.34} & 8.20 & 34.46\\
\bottomrule
\end{tabular}
\end{small}
\end{center}
\vskip -0.2in
\end{table}

\subsection{Low Pass Filter + Conv Kernel}
\begin{table}[t]
\caption{Comparison between convolution kernels w/ and w/o the LPF on the Charades dataset. Per-frame mAP is used as the evaluation metric. }
\label{tab:tlpm}
\vskip 0.1in
\begin{center}
\begin{small}
\begin{tabular}{lcccr}
\toprule
Method & Result\\
\midrule
I3D + super-events (baseline) & 18.4\% \\
I3D + TLPM + super-events & 22.9\% \\
I3D + truncated TLPM + super-events & \textbf{23.1\%} \\
\bottomrule
\end{tabular}
\end{small}
\end{center}
\vskip -0.15in
\end{table}

\begin{table}[t]
\caption{Comparison with state-of-the-art methods on Charades. Per-frame mAP is used as an evaluation metric.}
\label{tab:charades}
\vskip 0.15in
\begin{center}
\begin{small}
\begin{tabular}{lc}
\toprule
Method & Result (\%) \\
\midrule
R-C3D \cite{Xu2017} & 12.7 \\
SSN \cite{Zhao2017}& 16.4 \\
I3D \cite{Carreira2017} & 17.2 \\
TGM \cite{Piergiovanni2019a}& 22.3 \\
\midrule
Truncated TLPM & \textbf{23.1} \\
\bottomrule
\end{tabular}
\end{small}
\end{center}
\vskip -0.2in
\end{table}
\begin{table*}[t]
\caption{Temporal modeling performance between w/ and w/o LPF (Res50 backbone).}
\label{tab:recognition}
\vskip 0.15in
\begin{center}
\begin{small}
\begin{tabular}{lcccccc}
\toprule
\textbf{Dataset} & \textbf{Model} & \textbf{\#Frame} & \textbf{Resolution} & \textbf{Val Top-1} & \textbf{Val Top-5}\\
\midrule
\multirow{8}{*}{Something-Something V1}
&TSM & 8 & center crop & 45.62 & 74.24 \\
&TSM + LPF & 8 & center crop & 45.83 & 74.74 \\
&TSM & 8 $\times$ 2clip & full & 47.19 & 75.91 \\
&TSM + LPF & 8 $\times$ 2clip & full & 47.42 & 76.32 \\
&TSM & 16 & center crop & 47.11 & 77.04 \\
&TSM + LPF & 16 & center crop & 47.92 & 77.97 \\
&TSM & 16 $\times$ 2clip & full & 48.38 & 78.12 \\
&TSM + LPF & 16 $\times$ 2clip & full & 48.85 & 78.57 \\
\midrule
\multirow{8}{*}{Something-Something V2}
&TSM & 8 & center crop & 58.85 & 85.48 \\
&TSM + LPF & 8 & center crop & 59.78 & 85.95 \\
&TSM & 8 $\times$ 2clip & full & 61.24 & 87.13 \\
&TSM + LPF & 8 $\times$ 2clip & full & 61.63 & 87.50 \\
&TSM & 16 & center crop & 61.37 & 86.99 \\
&TSM + LPF & 16 & center crop & 61.87 & 87.29 \\
&TSM & 16 $\times$ 2clip & full & 63.09 & 88.16 \\
&TSM + LPF & 16 $\times$ 2clip & full & 63.23 & 88.46 \\
\midrule
\multirow{4}{*}{Kinetics 400}
&TSM & 8 & center crop & 70.92 & 89.93 \\
&TSM + LPF & 8 & center crop & 71.54 & 90.07 \\
&TSM & 8 $\times$ 10clip & center crop & 72.74 & 90.85 \\
&TSM + LPF & 8 $\times$ 10clip & center crop & 73.06 & 90.89 \\
\bottomrule
\end{tabular}
\end{small}
\end{center}
\vskip -0.15in
\end{table*}
In consideration of the 1D temporal convolution kernel can also be taken as a downsampling operation, LPF is inserted into the convolution operation to further verify the combination of LPF and downsampling operation. Following the TGM \cite{Piergiovanni2019a}, combine the LPF with convolution kernels as (TLPM):
\begin{equation}
K^{lp}= \sum_{p=1}^{P} w^a_p \cdot w^{lp}_{p},
\end{equation}
where $w^a$ represents the soft attention used to mix $P$ sets of LPF weights up, and $w^{lp}$ represents the low pass weight as defined in Equation~\ref{eqa:lpw}. The TLPM and the TGM have the same number of parameters. The TGM learns location parameter $\mu$ and width parameter $\sigma$ while the TLPM learns location parameter $\mu$ and cutoff frequency parameter $f_C$. The Gaussian function in TGM has rough boundaries according to the three-sigma rule, but the low pass function of TLPM is unbounded. Therefore, this paper truncates the low pass weight with an additional parameter $L$ as follows:
\begin{equation}
w^{tlp}_{n} = \left\{
\begin{aligned}
\sum_{m=1}^{N}w^{lp}_{n-m} & , &\vert n-m-\mu \vert \leq L, \\
0 \quad \quad \quad & , &\vert n-m-\mu \vert > L.
\end{aligned}
\right.
\end{equation}

\textbf{Implementation details.}
This section takes the best-performed model in \cite{Piergiovanni2018} which consists of 3 super-events layers as a baseline. After \cite{Piergiovanni2019a}, the TLPM layers are paired with 16 sets of low pass weights. Each of them with length 30 and c\_out 8. The model is trained with I3D features as input for 50 epochs using the Adam optimizer with a learning rate of 0.1 and a batch size of 16.

\textbf{Dataset.}
Charades dataset contains a total of 9848 videos among which 7986 are training videos and 1864 are testing videos. These videos are mostly about daily life and act according to provided scripts. Each video contains approximately 7 actions from 147 action classes. Different from THUMOS'14 and ActivityNet 1.3, one frame in Charades dataset might have multiple class labels to descript activities that happen at the same time, such as holding a cup of something and sitting in a chair. Per-frame mAP is used to measure the localization performances for the Charades dataset. All experiment results are evaluated using the official Charades\_v1\_localize file.

\textbf{Experimental results.} To make comparisons between the temporal convolution kernels with and without low pass filters, the baseline and the TLPM are evaluated on the Charades dataset. From Table~\ref{tab:tlpm} one can see that the TLPM outperforms the baseline by 4.5\% per-frame mAP. This demonstrates that the LPF is capable of alleviating the aliasing problem in convolution kernels, and anti-aliasing can improve the detection performance. Additionally, the truncated TLPM is tested to inspect the influence of boundaries. As shown in Table~\ref{tab:tlpm}, the truncated TLPM achieves a per-frame mAP of 23.1\% which outperforms the TLPM by 0.2\%. This result suggests that it is necessary and beneficial to restrict the range of the low pass weights following the TGM settings. The truncated TLPM is tested against other state-of-the-art models as can be seen from Table~\ref{tab:charades}. It shows competitive results and outperforms the state-of-the-art model TGM by 0.8\%.

\subsection{Extending to Temporal Modeling}
Extensive experiments were conducted on something-something \cite{Goyal2017} and Kinetics \cite{Kay2017} datasets to explore the utility of anti-aliasing with LPF in temporal modeling. In this section, TSM \cite{Lin2019a} was set as a baseline. Considering the consensus module using downsampling operations to merge segment-level temporal information, LPF was put before the consensus module. The implementation details followed subsection~\ref{section:sampling}, except for the changing of the cnn block into (conv1d(num\_class, num\_class, 3), relu, avgpool1d(num\_segment), conv1d(num\_class, num\_class, 1), sigmoid). The cnn block takes the segment features with shape (B, C, num\_segment) as input and outputs the shape (B, num\_class, 1). This means each channel of a video corresponds to an LPF. As shown in Table~\ref{tab:recognition}, TSM + LPF outperforms TSM consistently under various conditions, such as a different dataset, frame, and resolution settings. These results validate the effectiveness of anti-aliasing with LPF in temporal modeling preliminarily. Hopefully future works can investigate LPF in other temporal modeling models to provide more sources of evidence.

\section{Conclusion}
This paper proved that temporal downsampling operations with inadequate sampling rates usually lead to aliasing errors with traditional TAL methods. In this paper, low pass filters were used to alleviate the aliasing problem by inhibiting the high-frequency band. To maximize the benefits of anti-aliasing and retaining high-frequency information, the cutoff frequencies for different instances had to be adjusted dynamically. Extensive experimentation shows that anti-aliasing can promote the features' distinguishability, and properly eliminating part of high-frequency information for anti-aliasing is beneficial for the detection performance.

{\small
\bibliographystyle{ieee_fullname}
\bibliography{egbib}
}

\end{document}